\documentclass{article} % For LaTeX2e

\usepackage{iclr2021_conference,times}

\usepackage{amsmath,amsfonts,bm}

\def\eqref#1{equation~\ref{#1}}
\def\1{\bm{1}}

\DeclareMathAlphabet{\mathsfit}{\encodingdefault}{\sfdefault}{m}{sl}
\SetMathAlphabet{\mathsfit}{bold}{\encodingdefault}{\sfdefault}{bx}{n}

\def\sS{{\mathbb{S}}}

\newcommand{\KL}{D_{\mathrm{KL}}}

\usepackage{hyperref}
\usepackage{url}

\usepackage{booktabs}
\usepackage{multirow}
\usepackage{colortbl}
\usepackage{graphicx}
\usepackage{caption}
\usepackage{subcaption}
\usepackage{ textcomp }

\usepackage{xcolor}
\usepackage{soul}

\title{CO2: Consistent Contrast for Unsupervised Visual Representation Learning}

\iclrfinalcopy

\author{Chen Wei, Huiyu Wang, Wei Shen, Alan Yuille \\
Johns Hopkins University\\
\texttt{\{weichen3012,williamwanghuiyu,shenwei1231,alan.l.yuille\}@gmail.com}
}

\begin{document}

\maketitle

\begin{abstract}

Contrastive learning has been adopted as a core method for unsupervised visual representation learning. Without human annotation, the common practice is to perform an instance discrimination task: Given a query image crop, this task labels crops from the same image as positives, and crops from other randomly sampled images as negatives. An important limitation of this label assignment strategy is that it can not reflect the heterogeneous similarity between the query crop and each crop from other images, taking them as equally negative, while some of them may even belong to the same semantic class as the query. To address this issue, inspired by consistency regularization in semi-supervised learning on unlabeled data, we propose Consistent Contrast (CO2), which introduces a consistency regularization term into the current contrastive learning framework. Regarding the similarity of the query crop to each crop from other images as ``unlabeled'', the consistency term takes the corresponding similarity of a positive crop as a pseudo label, and encourages consistency between these two similarities. Empirically, CO2 improves Momentum Contrast (MoCo) by 2.9\% top-1 accuracy on ImageNet linear protocol, 3.8\% and 1.1\% top-5 accuracy on 1\% and 10\% labeled semi-supervised settings. It also transfers to image classification, object detection, and semantic segmentation on PASCAL VOC. This shows that CO2 learns better visual representations for these downstream tasks.

\end{abstract}

\section{Introduction}

%Unsupervised visual representation learning have recently attracted increasing research interests. By learning one or more pretext tasks without human labeling, models exploit the inherent knowledge within unlabeled visual data and learn features that are required to solve the pre-text tasks. These learned features have been shown useful, to some extent, for solving various downstream tasks. In this way, unsupervised visual representation learning unlocks the potential in large-scale pre-training for vision models without human annotation.

Unsupervised visual representation learning has attracted increasing research interests for it unlocks the potential of large-scale pre-training for vision models without human annotation. Most of recent works learn representations through one or more pretext tasks, in which labels are automatically generated from image data itself. Several early methods propose pretext tasks that explore the inherent structures within a single image. For example, by identifying spatial arrangement~\citep{contextprediction}, orientation~\citep{rotation}, or chromatic channels~\citep{colorization}, models learn useful representations for downstream tasks. Recently, another line of works~\citep{instdisc,amdim,cpc,cmc,moco,pirl,simclr}, e.g. Momentum Contrast (MoCo), falls within the framework of contrastive learning~\citep{contrasiveloss}, which directly learns relations of images as the pretext task. In practice, contrastive learning methods show better generalization in downstream tasks.

Although designed differently, most contrastive learning methods perform an instance discrimination task, i.e., contrasting between image instances. Specifically, given a query crop from one image, a positive sample is an image crop from the same image; negative samples are crops randomly sampled from other images in the training set. Thus, the label for instance discrimination is a one-hot encoding over the positive and negative samples. This objective is to bring together crops from the same image and keep away crops from different images in the feature space, forming an instance discrimination task.

However, the one-hot label used by instance discrimination might be problematic, since it takes all the crops from other images as equally negative, which cannot reflect the heterogeneous similarities between the query crop and each of them. For example, some ``negative'' samples are semantically similar to the query, or even belong to the same semantic class as the query. This is referred to as ``class collision''  in~\citet{theoretical} and ``sampling bias'' in \citet{debiased}.
%\chen{The ignorance of heterogeneous similarities to the crops from other images impairs the quality of the assigned labels, and consequently raises an obstacle for contrastive methods to learn a good representation.}
The ignorance of the heterogeneous similarities between the query crop and the crops from other images can thus raise an obstacle for contrastive methods to learn a good representation. A recent work, supervised contrastive learning~\citep{supervisedcl}, fixes this problem by using human annotated class labels and achieves strong classification performance. However, in unsupervised representation learning, the human annotated class labels are unavailable, and thus it is more challenging to capture the similarities between crops.

In this paper, we propose to view this instance discrimination task from the perspective of semi-supervised learning. The positive crop should be similar to the query for sure since they are from the same image, and thus can be viewed as labeled. On the contrary, the similarity between the query and each crop from other images is unknown, or unlabeled. With the viewpoint of semi-supervised learning, we introduce Consistent Contrast (CO2), a consistency regularization method which fits into current contrastive learning framework. Consistency regularization~\citep{cr} is at the core of many state-of-the-art semi-supervised learning algorithms~\citep{uda,mixmatch,fixmatch}. It generates pseudo labels for unlabeled data by relying on the assumption that a good model should output similar predictions on perturbed versions of the same image. Similarly, in unsupervised contrastive learning, since the query crop and the positive crop naturally form two perturbed versions of the same image, we encourage them to have consistent similarities to each crop from other images. Specifically, the similarity of the positive sample predicted by the model is taken as a pseudo label for that of the query crop.

Our model is trained with both the original instance discrimination loss term and the introduced consistency regularization term. %{\wsr{our consistency loss}{consistency regularization term}}
The instance discrimination label and the pseudo similarity label jointly construct a virtual soft label on-the-fly, and the soft label further guides the model itself in a bootstrap manner. %A semantically similar negative crop will no longer be pushed as far away as possible in the representation space as when only the one-hot label is used, but probably stays close thanks to the consistency term.
In this way, CO2 exploits the consistency assumption on unlabeled data, mitigates the ``class collision’’ effect introduced by the one-hot labels, and results in a better visual representation.

This simple modification brings consistent gains on various evaluation protocols. We first benchmark CO2 on ImageNet~\citep{imagenet} linear classification protocol. CO2 improves MoCo by $2.9\%$ on top-1 accuracy. It also provides 3.8\% and 1.1\% top-5 accuracy gains under the semi-supervised setting on ImageNet with 1\% and 10\% labels respectively, showing the effectiveness of the introduced consistency regularization. We also evaluate the transfer ability of the learned representations on three different downstream tasks: image classification, object detection and semantic segmentation. CO2 models consistently surpass their MoCo counterparts, showing that CO2 can improve the generalization ability of learned representation.

\section{Method}
In this section, we begin by formulating current unsupervised contrastive learning as an instance discrimination task. Then, we propose our consistency regularization term which addresses the ignorance of the heterogeneous similarity between the query crop and each crop of other images in the instance discrimination task.

\begin{figure}
\centering
\begin{subfigure}[b]{0.48\textwidth}
\centering
\includegraphics[width=\textwidth]{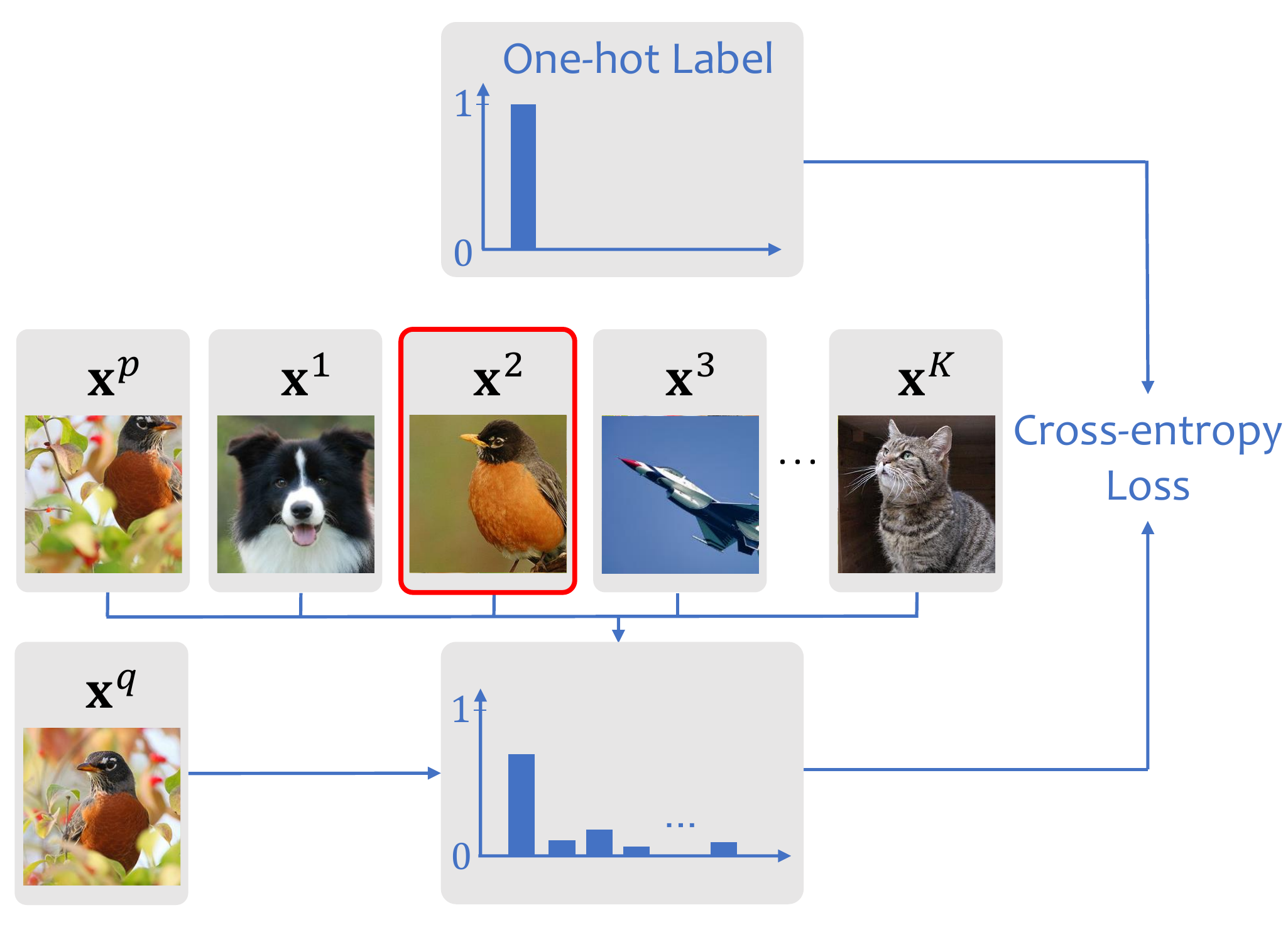}
\caption{Instance discrimination}
\label{fig:ia}
\vspace{-0.2cm}
\end{subfigure}
\hfill
\begin{subfigure}[b]{0.48\textwidth}
\centering
\includegraphics[width=\textwidth]{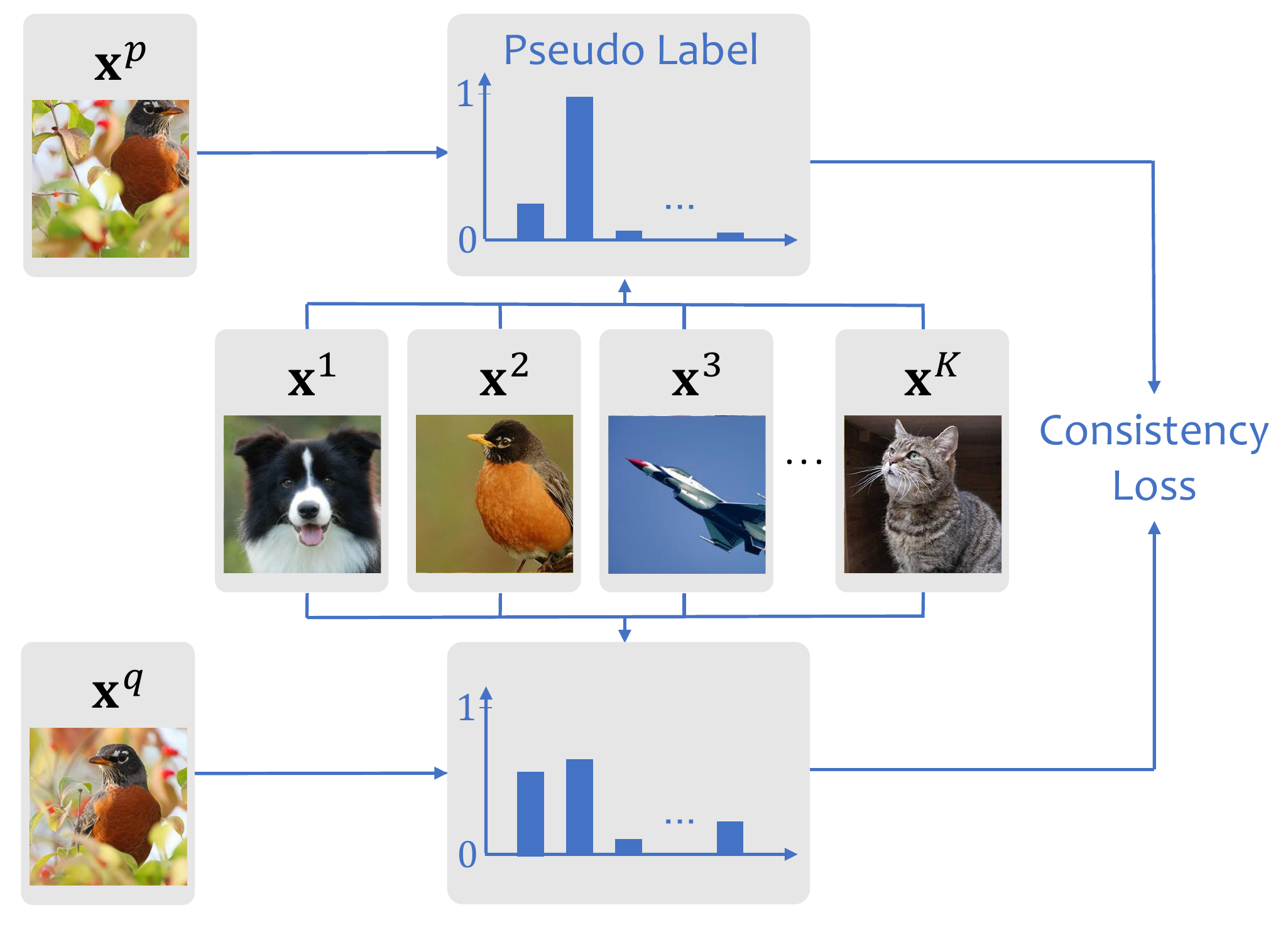}
\caption{Consistency regularization}
\label{fig:ib}
\vspace{-0.2cm}
\end{subfigure}
\label{fig:i}
\caption{Illustration of (\textbf{a}) instance discrimination and (\textbf{b}) our consistency regularization term. $\mathbf{x}^q$ represents the query crop, $\mathbf{x}^p$ represents a positive crop from the same image, and $\{\mathbf{x}^k\}_{k=1}^K$ represents $K$ negative crops from other images. In instance discrimination, the query is labeled by a one-hot encoding identifying the positive. However, some negatives can be semantically similar (red box) but are not reflected by the one-hot label. In consistency regularization, the similarity between the positive and the negatives are taken as a pseudo label and we encourage the agreement between the query and the positive.}
\vspace{-0.3cm}
\end{figure}

\subsection{Contrastive Learning}

Contrastive learning~\citep{contrasiveloss} is recently adopted as an objective for unsupervised learning of visual representations. Its goal is to find a parametric function $f_{\theta}:\mathbb{R}^D\rightarrow \mathbb{R}^d$ that maps an input vector $\mathbf{x}$ to a feature vector $f_{\theta}(\mathbf{x}) \in \mathbb{R}^d$ with $D\gg d$, such that a simple distance measure (e.g., cosine distance) in the low-dimensional feature space can reflect complex similarities in the high-dimensional input space.

For each input vector $\mathbf{x}_i$ in the training set $\sS$, the similarity measure in the input space is defined by a subset of training vectors $\sS_i \subset \sS$, called similarity set. The sample $\mathbf{x}_i$ is deemed similar to samples in the similarity set $\sS_i$, but dissimilar to samples in $\sS \setminus \sS_i$. Then, the contrastive objective encourages $f_{\theta}(\mathbf{x}_j)$ to be close to $f_{\theta}(\mathbf{x}_i)$ in the feature space if $\mathbf{x}_j \in \sS_i$, and otherwise to be distant.

By training with contrastive loss, the similarities defined by the similarity set determine characteristics of the learned representation and the mapping function $f_{\theta}$. For example, if the similarity is defined as samples from the same semantic class, then $f_{\theta}$ will probably learn invariances to other factors, e.g., object deformation. In the supervised setting, this definition of similarity requires a large amount of human labeling. On the contrary, unsupervised contrastive learning exploits similarities with no need of human labels. One natural definition of unsupervised similarity is multiple views of an image, as explored by many recent methods. For example, random augmented crops~\citep{instdisc, clusterspread, moco, simclr, mocov2} of an image could be defined as a similarity set. %Formally, each training image $I_i$ corresponds to one similarity set $\sS_i$, which consists of random augmented crops $\{\mathbf{x}_i^1, \mathbf{x}_i^2, \dots\}$ of image $I_i$.
In this case, the contrastive objective is effectively solving an instance discrimination task~\citep{instdisc} as illustrated in Figure~\ref{fig:ia}.

%\huiyu{x is not used in contrast, but in the first paragraph},
The training of this instance discriminator involves randomly sampling a query crop $\mathbf{x}^q \in \sS_i$, a positive crop $\mathbf{x}^p \in \sS_i$ from the same image, and $K$ negative crops $\{ \mathbf{x}^k \in \sS \setminus \sS_i \}_{k=1}^K$ from other images. These $K+2$ crops (the query, the positive, and $K$ negatives) are encoded with $f_\theta$ respectively, $\mathbf{q}=f_\theta(\mathbf{x}^q), \mathbf{p}=f_\theta(\mathbf{x}^p), \mathbf{n}_k=f_\theta(\mathbf{x}^k)$. Then, an effective contrastive loss function, InfoNCE~\citep{cpc}, is written as:
\begin{equation}
\mathcal{L}_{ins} = -\log \frac{\exp (\mathbf{q}\cdot \mathbf{p} / \tau_{ins})}{\exp (\mathbf{q}\cdot \mathbf{p} / \tau_{ins}) + \sum_{k=1}^{K} \exp(\mathbf{q}\cdot \mathbf{n}_k / \tau_{ins})}\,, 
\end{equation}
where $\tau_{ins}$ is a temperature hyper-parameter~\citep{knowledgedistillation}. This loss can be interpreted as a cross entropy loss that trains the model to discriminate the positive crop (labeled as $1$) from negative crops (labeled as $0$) given the query crop. We denote this loss as $\mathcal{L}_{ins}$ as it performs an instance discrimination task.

\subsection{Consistent Contrast}

The one-hot labels used by InfoNCE loss is effective, showing good generalization ability across tasks and datasets~\citep{mocov2, simclr}. Nevertheless, we argue that the hard, zero-one labels is uninformative. Specifically, crops from other images are taken as equally negative as they are all labeled as 0. This is contradictory to the fact that some so-called ``negative'' crops can be similar or even in the same semantic class, especially when $K$ is large. For example, SimCLR \citep{simclr} uses 16,382 negative samples in a batch, and MoCo \citep{moco, mocov2} uses a memory bank of 65,536 features as negative samples. Even worse, the current objective forces negatives to be as far from the query as possible, with larger weights for closer negatives since they are ``hard negatives''. However, these ``hard negative'' crops in fact tend to be semantically close. These issues impair good representation learning because the one-hot labels can not faithfully reflect the heterogeneous similarities between the query crop and the crops from other images.

Although generating labels based on instance discrimination is trivial, revealing the similarity between two arbitrary crops is exactly what we want to learn from unsupervised pre-training. Therefore, the label of the similarity between the query crop and each crop from other images is of little hope to get. This situation is similar to the usage of unlabeled data in semi-supervised learning setting, in which consistency regularization is widely used to propagate knowledge from labeled data to discover the structures in unlabeled data. Inspired by this, we propose to encourage the consistency between the similarities of crops from the same image, i.e., the query crop and the positive crop. We illustrate the consistency regularization in Figure~\ref{fig:ib}.

First, we denote the similarity between the query $\mathbf{q}$ and the negatives $\mathbf{n}_i (i\in\{1,\ldots,K\})$ as:
\begin{equation}
Q(i) = \frac{\exp(\mathbf{q} \cdot \mathbf{n}_i / \tau_{con})}{\sum_{k=1}^K \exp(\mathbf{q} \cdot \mathbf{n}_k / \tau_{con})}\,,
\end{equation}
where $\tau_{con}$ is also a temperature hyper-parameter. $Q(i)$ is the probability that the query $\mathbf{q}$ selects $\mathbf{n}_i$ as its match from $\{\mathbf{n}_k\}_{k=1}^K$. Similarly, the similarity between the positive $\mathbf{p}$ and the negatives is written as:
\begin{equation}
P(i) = \frac{\exp(\mathbf{p} \cdot \mathbf{n}_i / \tau_{con})}{\sum_{k=1}^K \exp(\mathbf{p} \cdot \mathbf{n}_k / \tau_{con})}\,.
\end{equation}
We impose the consistency between the probability distributions $P$ and $Q$ by using symmetric Kullback-Leibler (KL) Divergence as the measure of disagreement:
\begin{equation} \label{eq:KL}
\mathcal{L}_{con} = \frac{1}{2} \KL (P \Vert Q) + \frac{1}{2} \KL (Q \Vert P)\,.
\end{equation}
As shown in Equation~\ref{eq:KL}, the predicted similarity of the positive crop to each crop of the other images, i.e., $P$, are served as a soft pseudo label to those of the query, i.e., $Q$. The total loss is a weighted average of the original instance discrimination loss term and the consistency regularization term: 
\begin{equation}
\mathcal{L} = \mathcal{L}_{ins} + \alpha \mathcal{L}_{con}\,,
\end{equation}
where $\alpha$ denotes the coefficient to balance the two terms. It is possible to merge the two terms by creating a unique label containing information of both the one-hot label and the pseudo similarity label, but we find the weighted average can already get good performance and is easy to control.

%Why is $P$ a reasonable label for $Q$? First, the query crop and the positive crop are from the same instance $i$, so they share the same relative similarity. Second, even though not perfect, the embedding function $f$ has generalized to measure the relative similarity in the input space to some extend. In other words, crops from similar images tend to get higher scores and crops from dissimilar images tend to get lower scores in $P(j)$ according to the knowledge learned by the model. And $\mathcal{L}_{s}$ provides a way to bootstrap the learned knowledge of contrast between images. An alternative perspective of $\mathcal{L}_{s}$ is that it maximizes the consistency of similarity relationship based on two different views of the same image.

The pseudo label $P$ is informative to reveal the similarity between the query $\mathbf{q}$ and each $\mathbf{n}_i$, while the one-hot label is unable to provide such information, since it only describe co-occurrence within one image. Note that, the pseudo label $P$ is also dynamic since the embedding function $f_\theta$ is updated in every training step, and thus generating better pseudo labels during training. It indicates that the unsupervised embedding function and the soft similarity labels give positive feedback to each other.

%Wei {\wsr{used in $\mathcal{L}_{con}$}{}} is soft and dynamic. {\wsr{The soft label $P$}{it}} is informative to reveal the similarity between the query $\mathbf{q}$ and each $\mathbf{n}_i$, which can not be provided by the one-hot label describing only co-occurrence within one image. The pseudo label $P$ is dynamic since $f$ is updated in every training step, generating better labels during training. It suggests that the unsupervised embedding function can refine the soft similarity labels while refining the embedding itself.

Our method is simple and low-cost. It captures the similarity to each $\mathbf{n}_i$ while introducing unnoticeable computational overhead with only one extra loss term computed. This is unlike clustering based unsupervised learning methods, which are costly, since they explicitly compute the similarity sets in the training set after every training epoch~\citep{deepclustering, la, pcl}.

%Wei other methods utilizing unsupervised clustering such as k-means to explicitly model the similar sets in the training set after every training epoch~\citep{deepclustering, la, pcl}, which is costly.

% Add "results" to each task: linear eval, semi-supervised, transfer. e.g. improves from 60.6% to 63.5%. The current version looks like the "setup" subsection in the MoCo paper.
% Some technical details can be skipped to save space. e.g. we follow MoCo.

\section{Experiments}

Herein, we first report our implementation details and benchmark the learned representations on ImageNet. Next, we examine how the unsupervised pre-trained models transfer to other datasets and tasks. We then analyze the characteristics of our proposed method.

%We mainly compare our model with both MoCo and its extension MoCo v2~\citep{mocov2} which utilizes cosine learning rate schedule, extra blur augmentation and an MLP head.  our Learned Contrast  is orthogonal to these advanced designs. \huiyu{move to "Technical details"?}

\subsection{Implementation}

\begin{table}
  \caption{Linear classification protocol on ImageNet-1K}
  \vspace{-0.2cm}
  \label{tab:linear-cls}
  \centering
  \begin{tabular}{llccc}
    \toprule
    Pretext Task              & Arch.  & Head    & \#epochs & Top-1 Acc. (\%)         \\
    \midrule
    ImageNet Classification             & R50           & -       & 90       &     { }76.5      \\
    \arrayrulecolor{black!10}\midrule
    Exemplar~\citep{exemplar}  & R50w$3\times$     & -       & 35       & { }46.0 \\
    Relative Position~\citep{contextprediction} & R50w$2\times$     & -       & 35       &  { }51.4 \\
    Rotation~\citep{rotation}  & Rv50w$4\times$    & -       & 35       & { }55.4 \\
    Jigsaw~\citep{jigsaw}      & R50               & -       & 90       & { }45.7 \\
%    Colorization~\citep {colorization} & R50        & -       & 28       &   { }39.6  \\
%    DeepCluster~\citep{deepclustering}& VGG        & -       & 100      & { }48.4 \\
%    BigBiGAN~\citep{bigbigan}  & R50               & -       & -        & { }56.6  \\
%    SelfLabel~\citep{selflabel} & R50              & -       & 400      &     { }61.5 \\
    \arrayrulecolor{black!100}\midrule
    \multicolumn{5}{l}{\emph{Methods based on contrastive learning}:} \\
    \midrule
    InsDisc~\citep{instdisc}  & R50               & Linear   & 200       &    { }54.0       \\
    Local Agg.~\citep{la}     & R50                 & Linear   & 200       &    { }58.2       \\
    CPC v2~\citep{cpcv2}      & R170$_\mathit{w}$         & -        & \texttildelow 200       & { }65.9  \\
    CMC~\citep{cmc}           & R50               & Liner    & 240       &  { }60.0 \\
    AMDIM~\citep{amdim}       & AMDIM$_\mathit{large}$            & -        & 150       &  { }68.1  \\
    PIRL~\citep{pirl}         & R50                & Linear   & 800       &  { }63.6          \\
    SimCLR~\citep{simclr}     & R50               & MLP      & 1000      &  { }69.3          \\
    \midrule
    MoCo~\citep{moco}      & R50              & Linear   & 200       &  { }60.6          \\
    MoCo~\citep{moco} + CO2 & R50           & Linear   & 200       &  { }63.5          \\
    MoCo v2~\citep{mocov2}    & R50               & MLP      & 200       &  {  }67.5          \\
    MoCo v2~\citep{mocov2} + CO2 & R50             & MLP      & 200       & { }68.0             \\
    \bottomrule
  \end{tabular}
  \flushleft
 %* \textit{Using Fast AutoAugment~\citep{fastauto}, which is supervised by ImageNet labels.}
\vspace{-0.2cm}
\end{table}

\paragraph{Setup} We evaluate CO2 based on MoCo~\citep{moco} and MoCo v2~\citep{mocov2}. Both of them use instance discrimination as pretext task, while MoCo v2 adopts more sophisticated design choices on projection head architecture, learning rate schedule and data augmentation strategy. We test CO2 on MoCo for its representativeness and simplicity. On MoCo v2, we evaluate how CO2 is compatible with advanced design choices. Note that although we evaluate CO2 within the framework of MoCo, it can be easily applied to other contrastive learning mechanisms.

The unsupervised training is performed on the train split of ImageNet-1K~\citep{imagenet} without using label information. We keep aligned every detail with our baseline MoCo to effectively pin-point the contribution of our approach, except the number of GPUs (MoCo uses 8 GPUs while we use 4). A further search on MoCo-related hyper-parameters might lead to better results of our method. For the hyper-parameters of CO2, we set $\tau_{con}$ as 0.04, $\alpha$ as 10 for MoCo-based CO2, and $\tau_{con}$ as 0.05, $\alpha$ as 0.3 for MoCo v2-based CO2. Please refer to the appendix for more detailed implementation description.

\begin{table}
  \caption{Top-5 accuracy for semi-supervised learning on ImageNet}
  \vspace{-0.2cm}
  \label{tab:semi-sup}
  \centering
  \begin{tabular}{lcc}
    \toprule
    Pretext Task                   & 1\% labels     & 10\% labels         \\
   \midrule
    Supervised Baseline                  &  48.4          &   80.4 \\
    \arrayrulecolor{black!10}\midrule
    InsDisc~\citep{instdisc}    &  39.2       &   77.4       \\
    PIRL~\citep{pirl}           &  57.2       & 83.8          \\
    \arrayrulecolor{black!100}\midrule
    MoCo~\citep{moco}      & 62.4       & 84.1          \\
    MoCo~\citep{moco} + CO2   & 66.2       & 85.2         \\
    MoCo v2~\citep{mocov2}     & 69.8       & 85.0          \\
    MoCo v2~\citep{mocov2} + CO2    & 71.0       & 85.7             \\
    \bottomrule
  \end{tabular}
\vspace{-0.2cm}
\end{table}
\subsection{Linear Classification}
We first benchmark the learned representations on the common linear classification protocol. After the unsupervised pre-training stage, we freeze the backbone network including the batch normalization parameters, and train a linear classifier consisting of a fully-connected layer and a softmax layer on the 2048-D features following the global average pooling layer. Table~\ref{tab:linear-cls} summaries the single-crop top-1 classification accuracy on the validation set of ImageNet-1K. Our method consistently improves by 2.9\% on MoCo and by 0.5\% on MoCo v2. We also list several top-performing methods in the table for reference. These results indicate that the representation is more linearly separable on ImageNet with consistency regularization, since the consistency regularization mitigates the ``class collision'' effect caused by semantically similar negative samples.

\subsection{Semi-supervised Learning}
We next perform semi-supervised learning on ImageNet to evaluate the effectiveness of the pre-trained network in data-efficient settings. Following~\citep{instdisc, pirl, simclr}, we finetune the whole pre-trained networks with only 1\% and 10\% labels which are sampled in a class-balanced way. Table~\ref{tab:semi-sup} summaries the top-5 accuracy on the validation set of ImageNet-1K. The results for MoCo and MoCo v2 are produced by us using their officially released models. The proposed consistency regularization term can provide 3.8\% and 1.2\% top-5 accuracy gains for MoCo with 1\% and 10\% labels respectively. CO2 also improves from MoCo v2 by 1.2\% top-5 accuracy with 1\% labels, and by 0.7\% with 10\% labels.

\begin{table}
  \caption{Transfer learning performance on PASCAL VOC datasets}
  \vspace{-0.2cm}
  \label{Ttab:transfer}
  \centering
  \begin{tabular}{lcrclc}
    \toprule
    
                & Image  &  \multicolumn{3}{c}{Object} & Semantic \\
                & Classification &  \multicolumn{3}{c}{Detection} & Segmentation \\
    \cmidrule(r){2-6}
    Pretext Task &  mAP   & {AP}\textsubscript{50}\quad   & {AP}\textsubscript{all}\quad     & AP\textsubscript{75}    &  mIoU  \\
    \midrule
    ImageNet Classification & 88.0 & 81.3   & 53.5   & 58.8 & 74.4  \\
    \arrayrulecolor{black!10}\midrule
    Rotation~\citep{rotation} & 63.9  & 72.5   & 46.3   & 49.3 & -    \\
    Jigsaw~\citep{jigsaw}     & 64.5& 75.1   & 48.9   & 52.9 & -     \\
    InsDisc~\citep{instdisc}  & 76.6  & 79.1   & 52.3   & 56.9 & -     \\ 
    PIRL~\citep{pirl}         & 81.1  & 80.7   & 54.0   & 59.7 & -     \\
    \arrayrulecolor{black!100}\midrule
    MoCo~\citep{moco}             & -   & 81.5   & 55.9   & 62.6    & 72.5  \\
    MoCo~\citep{moco} (\textit{our impl.}) & 79.7  & 81.6   & 56.2   & 62.4      & 72.6  \\
    MoCo~\citep{moco} + CO2      & 82.6    & 81.9   & 56.0  & 62.6    & 73.3     \\
    MoCo v2~\citep{mocov2}   & 85.0     & 82.4   & 57.0   & 63.6   & 74.2  \\
    MoCo v2~\citep{mocov2} + CO2      & 85.2     &82.7   & 57.2  & 64.1    & 74.7     \\

    \bottomrule
  \end{tabular}
\vspace{-0.2cm}
\end{table}

\subsection{Transfer Learning}

To further investigate the generalization ability of our models across different datasets and tasks, we evaluate the transfer learning performance on PASCAL VOC~\citep{pascal} with three typical visual recognition tasks, i.e., image classification, object detection and semantic segmentation. Table~\ref{Ttab:transfer} reports the transfer learning performance comparing with other methods using ResNet-50. Our proposed CO2 shows competitive or better performance comparing with the corresponding baselines.
\vspace{-0.3cm}
\paragraph{Image Classification} Following the evaluation setup in \citet{fairssl}, we train a linear SVM~\citep{svm} on the frozen 2048-D features extracted after the global average pooling layer. The results of MoCo are produced by us with their official models. In this case, CO2 is 2.9\% better than MoCo, and 0.2\% than MoCo v2.
\vspace{-0.3cm}
\paragraph{Object Detection} Following the detection benchmark set up in \citet{moco}, we use Faster R-CNN~\citep{fasterrcnn} object detector and ResNet-50 C4~\citep{maskrcnn} backbone, and all the layers are finetuned including the batch normalization parameters. The numbers of our method are averaged over three runs. Our reproduced results for MoCo are also listed in the table for reference. CO2 provides 0.3\% AP\textsubscript{50} gains on both MoCo and MoCo v2.
\vspace{-0.3cm}
\paragraph{Semantic Segmentation} We follow the settings in \citet{moco} for semantic segmentation. Results are average over three runs. Similarly, we include our reproduced results of MoCo as a reference. The result of MoCo v2 is produced by us using its officially released model. CO2 gives 0.9\% mIoU improvement upon MoCo, and 0.5\% upon MoCo v2, which finally surpasses its supervised counterpart.

%\begin{table}
%  \caption{Ablation}
%  \label{ablation}
%  \centering
%  \begin{tabular}{lc}
%    \toprule
    
%    Ablation & Accuracy (\%)  \\
%    \midrule
%    end-to-end        & - \\
%    end-to-end + ours & - \\
%    moco v1 + ours with MSE loss & - \\
%    moco v1 + label smoothing & - \\
%    moco v1 + ours with KL loss  & - \\
%    moco v1 + ours with JSD loss & - \\
%    moco v1 + ours with T = 0.1  & - \\
%    moco v1 + ours with T = 0.01 & - \\
%    \bottomrule
%  \end{tabular}
%\end{table}

\subsection{Analysis}
\label{sec:analysis}

In this section, we study the characteristics of the proposed method on a smaller backbone ResNet-18 and a smaller dataset ImageNet-100 due to the consideration of the computational resource. ImageNet-100 is firstly used in \citet{cmc} and consists of 100 randomly selected classes from all $1,000$ classes of ImageNet.

\begin{figure}
\centering
\begin{subfigure}[b]{0.47\textwidth}
\centering
\includegraphics[width=\textwidth]{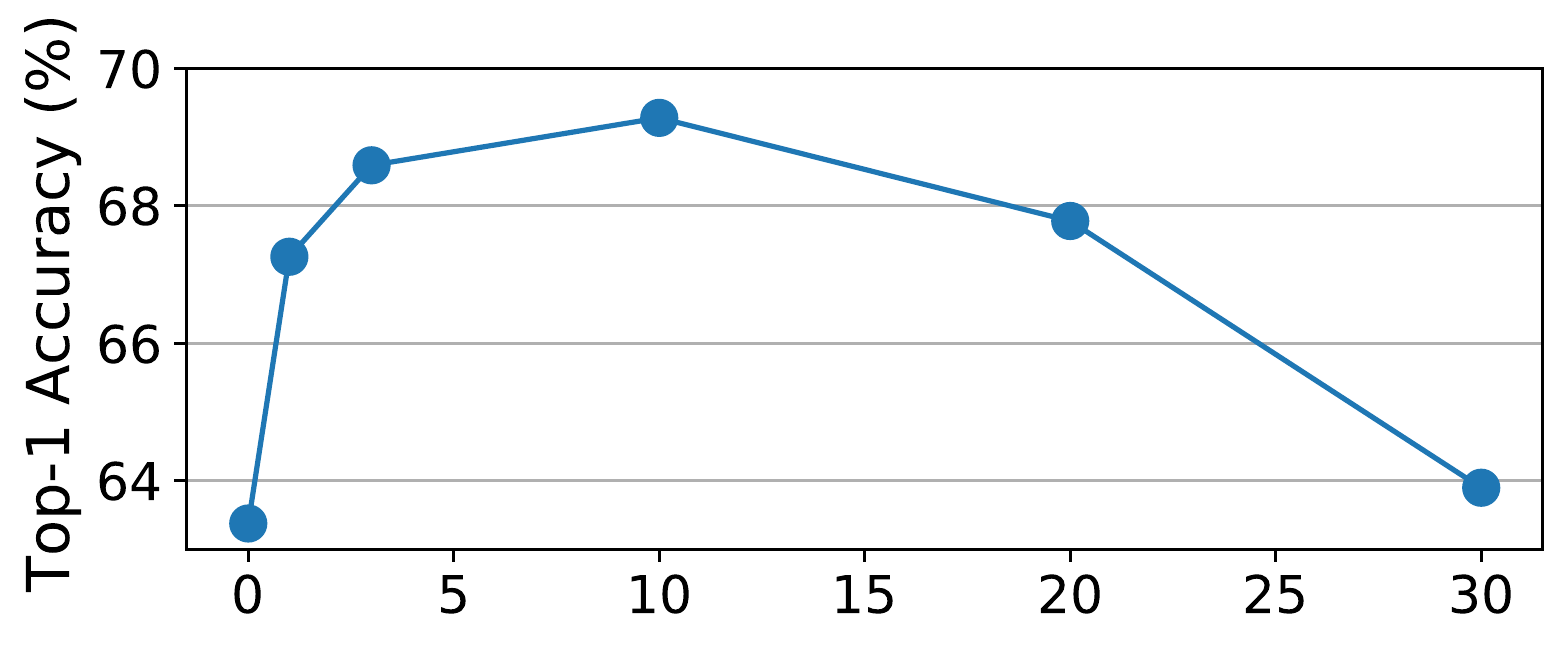}
\caption{Effect of varying the coefficient $\alpha$.}
\label{fig:hpa}
\end{subfigure}
\hfill
\begin{subfigure}[b]{0.47\textwidth}
\centering
\includegraphics[width=\textwidth]{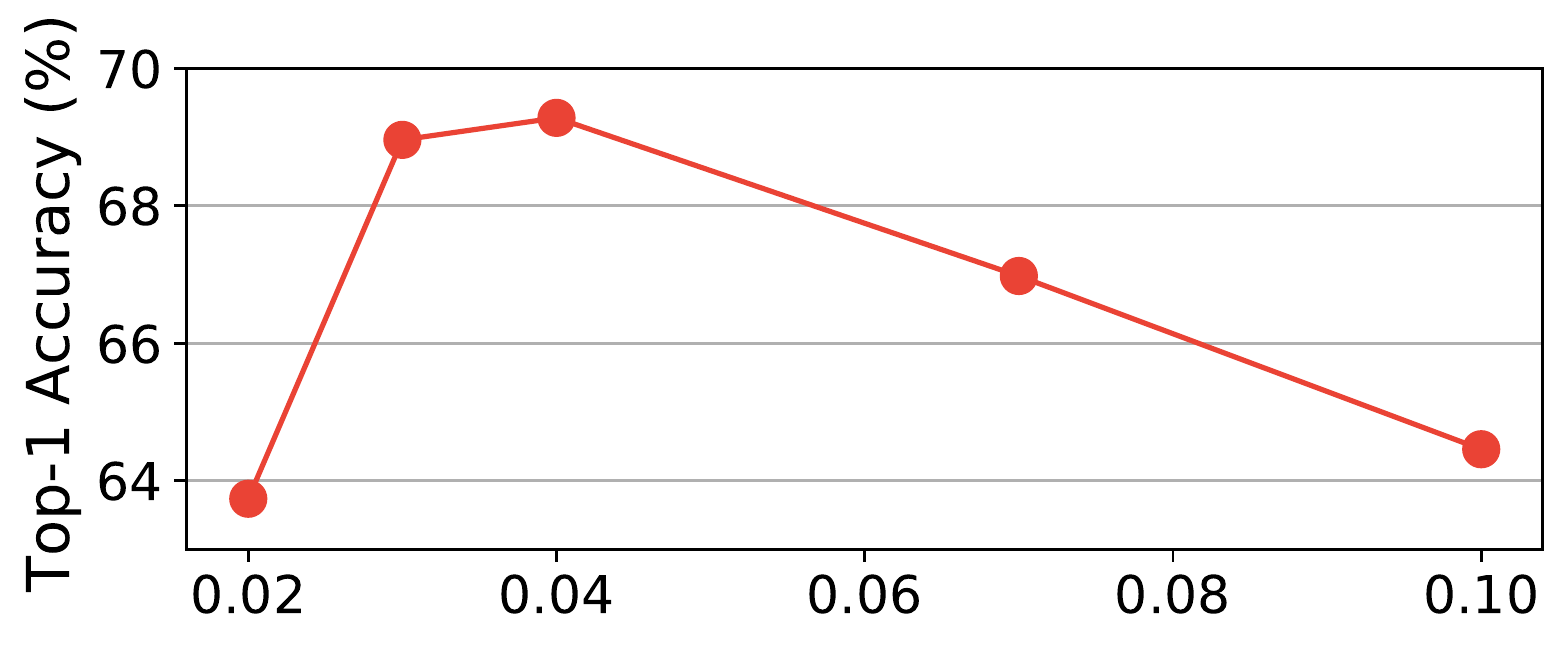}
\caption{Effect of varying the temperature $\tau_{con}$.}
\label{fig:hpt}
\end{subfigure}
\caption{Ablation on the effect of hyper-parameters.}
\label{fig:hp}
\vspace{-0.2cm}
\end{figure}

\begin{figure}
\centering
\begin{subfigure}[b]{0.31\textwidth}
\centering
\includegraphics[width=\textwidth]{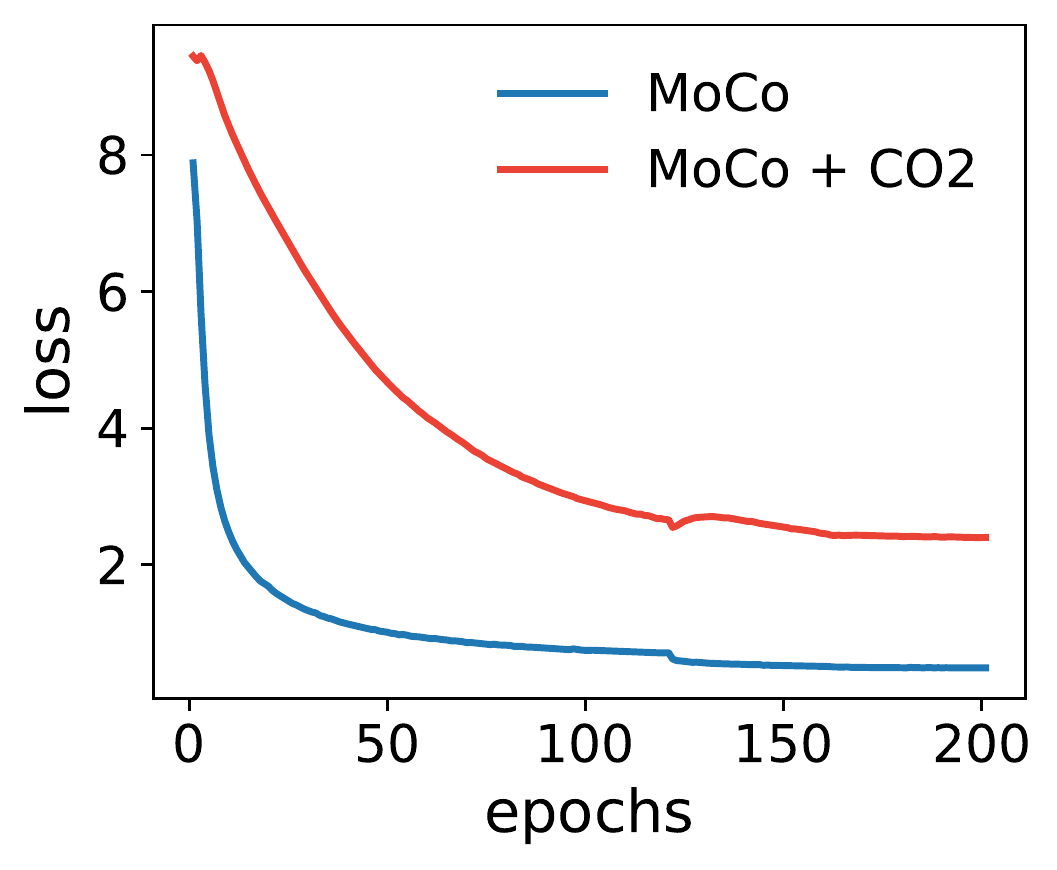}
\caption{$\mathcal{L}_{ins}$}
\label{fig:csm}
\end{subfigure}
\hfill
\begin{subfigure}[b]{0.31\textwidth}
\centering
\includegraphics[width=\textwidth]{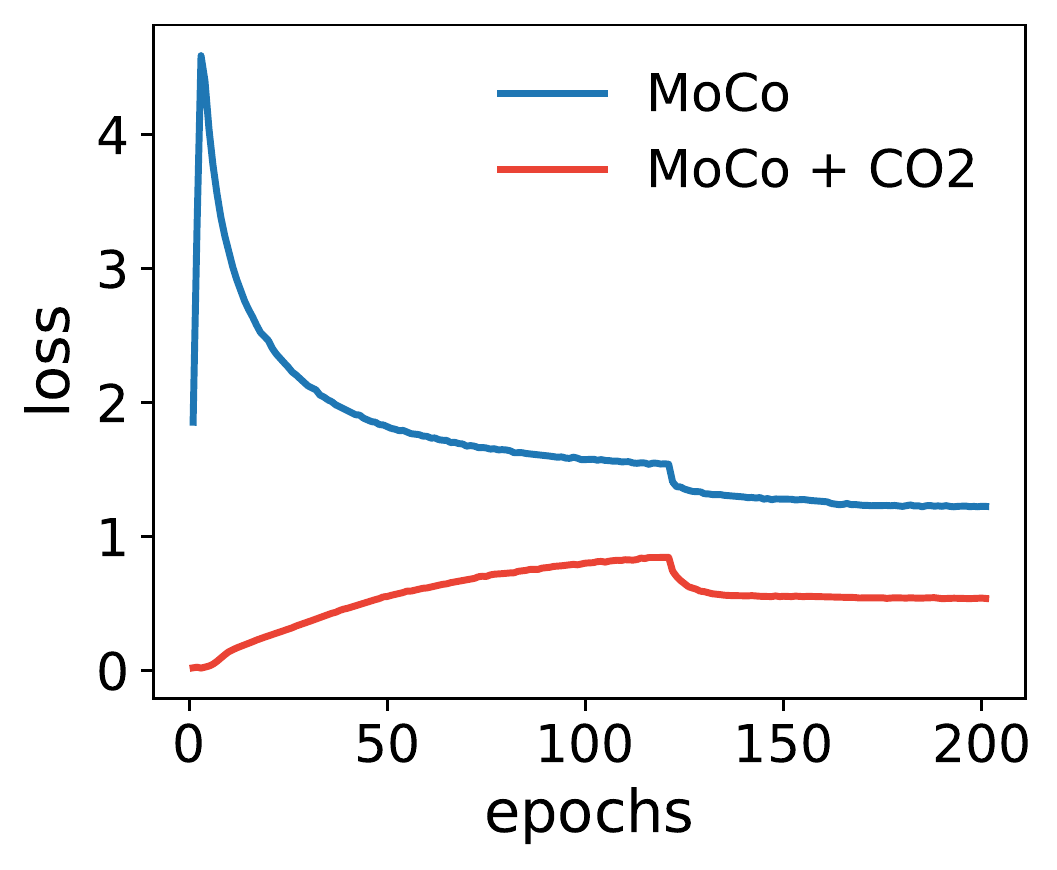}
\caption{$\mathcal{L}_{con}$}
\label{fig:ckl}
\end{subfigure}
\hfill
\begin{subfigure}[b]{0.325\textwidth}
\centering
\includegraphics[width=\textwidth]{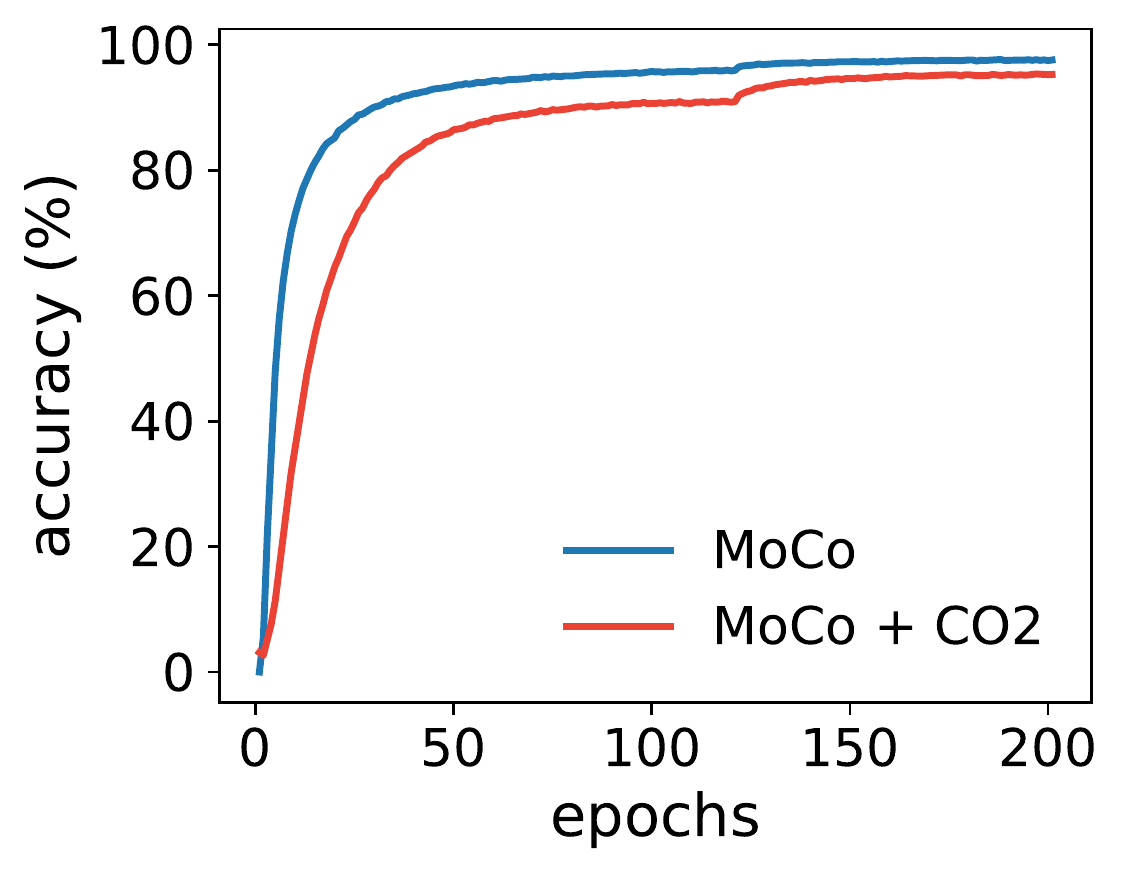}
\caption{Instance discrimination acc. }
\label{fig:cacc}
\end{subfigure}
\caption{Training curves of ResNet-18 on ImageNet-100.}
\label{fig:c}
\vspace{-0.3cm}
\end{figure}

\vspace{-0.3cm}
\paragraph{Hyper-parameter} Our method introduces two new hyper-parameters, the coefficient of consistency regularization term $\alpha$, and its temperature $\tau_{con}$. In Figure~\ref{fig:hp}, we show the top-1 accuracy of a linear classifier on models pre-trained by CO2 with different hyper-parameters. In Figure~\ref{fig:hpa}, we fix the temperature $\tau_{con}$ as 0.04 and vary the coefficient $\alpha$. The best coefficient is 10. We see that by using the consistency regularization term, the linear classification accuracy can be boosted from 63.6\% to 69.2\%. Increasing $\alpha$ to 20 and beyond causes performance degeneration. We hypothesize that the model is over-regularized by the consistency loss, and thus it loses some discrimination among different instances. In Figure~\ref{fig:hpt}, we fix the coefficient to be 10 and varying the temperature. As other consistency regularization methods (e.g., \citet{mixmatch}), temperature $\tau_{con}$ effectively influences the quality of the learned representation, and the best to use is 0.04. 
\vspace{-0.3cm}
\paragraph{Training Curves} In Figure~\ref{fig:c} we show the training curves of the instance discrimination loss $\mathcal{L}_{ins}$, the consistency loss $\mathcal{L}_{con}$ and the instance discrimination accuracy. Instance discrimination accuracy represents the percent of query crops which successfully select their corresponding positive crops, i.e., successfully identify their instances. MoCo is trained with $\mathcal{L}_{ins}$ only and its $\mathcal{L}_{con}$ is just calculated out for comparison. We see that $\mathcal{L}_{ins}$ of MoCo drops quickly from the beginning at the cost of a jump of $\mathcal{L}_{con}$. As the training proceeds, $\mathcal{L}_{con}$ of MoCo decreases spontaneously, possibly because more semantic knowledge has been learned, but it is still relatively high. Training with $\mathcal{L}_{con}$ and $\mathcal{L}_{ins}$ together, i.e., MoCo + CO2, $\mathcal{L}_{con}$ is kept very low from beginning, and $\mathcal{L}_{con}$ increases gradually since the model is trained to discriminate between images at the same time. At the end of the training, $\mathcal{L}_{con}$ stays much lower than $\mathcal{L}_{con}$ of MoCo.

We also notice that with CO2, the instance discrimination accuracy drops from 97.57\% to 95.26\%. Although CO2 results in lower instance discrimination accuracy, it still does better in the downstream classification task. The linear classification accuracy improves from 63.6\% to 69.2\%, as shown in Figure~\ref{fig:hpa}. It suggests again that there is a gap between instance discrimination and the downstream tasks. %which are of really interest.   %When optimized with $\mathcal{L}_{con}$, the consistency regularization term, $\mathcal{L}_{con}$ is significantly lower than without $\mathcal{L}_{con}$. On the contrary, the instance discrimination loss $\mathcal{L}_{ins}$ is higher and the instance discrimination accuracy is lower. This shows that the consistency regularization can effectively slow down the model from overfitting to the instance discrimination task but not learning much semantic knowledge.
\vspace{-0.3cm}
\paragraph{Comparision with Label Smoothing} With the consistency regularization term, our approach assigns soft pseudo labels to crops from other images. This looks similar to label smoothing regularization on supervised classification~\citep{ls}, a useful trick which assigns a small constant value to the labels of all the negative classes to avoid overconfidence. We equip MoCo with label smoothing, i.e.,  assigning a small constant value to crops from other images (the ``negatives''). Surprisingly, it reports 61.2\% linear classification accuracy, 2.4\% lower than MoCo alone. This suggests that assigning a constant value as label smoothing can be harmful for unsupervised contrastive learning, since it ignores the heterogeneous similarity relationship. And it is better to assign labels according to the similarities as our consistency regularization.

%{\wsr{which we hypothesis is because that the model does not learn a good enough starting point in the first place.}{We hypothesize that the reason might be the model is over-regularized by the consistency loss, and thus it loses some discrimination among different instances.}} In Figure~\ref{fig:hpt}, we fix the coefficient to be $10$ and varying the temperature. As other InfoNCE based contrastive learning methods, temperature $\tau_s$ effectively influences the quality of the learned representations, and we find the best temperature to use is 0.04. 

%\paragraph{Comparison with Label Smoothing}

\section{Related Work}

Our method falls in the area of unsupervised visual representation learning, especially for image data. In this section, we first revisit various design strategies of pretext tasks for unsupervised learning. Then we elaborate on the pretext tasks based on contrastive learning, which is the focus of our work. Next, we review the methods using consistency regularization in semi-supervised learning, which inspire our work. 
\vspace{-0.3cm}
\paragraph{Unsupervised Learning and Pretext Tasks} To learn from unlabeled image data, a wide range of pretext tasks have been established. The models can be taught to specify the relative position of a patch~\citep{contextprediction}, solve spatial jigsaw puzzles~\citep{jigsaw,arbitrarypuzzle}, colorize gray scale images~\citep{colorization,colorizationproxy}, inpaint images~\citep{inpainting}, count objects~\citep{counting}, discriminate orientation~\citep{rotation}, iteratively cluster~\citep{deepclustering,la,selflabel,deeprobustclustering}, generate images~\citep{bigan, bigbigan}, \textit{etc}. \citet{multi} evaluates the combination of different pretext tasks. \citet{revisting} and \citet{fairssl} revisit and benchmark different pretext tasks.
\vspace{-0.3cm}
\paragraph{Contrastive Learning} Contrastive learning~\citep{contrasiveloss} recently puts a new perspective on the design of pretext task and holds the key to most state-of-the-art methods. Most of them perform an instance discrimination task while differ in i) the strategies to synthesize positives and negatives, and ii) the mechanisms to manage a large amount of negatives. The synthesizing can base on context with patches~\citep{cpc,dim}, random resized crops with data augmentation~\citep{instdisc,clusterspread,amdim,moco,simclr}, jigsaw puzzle transformation~\citep{pirl} or luminance-chrominance decomposition~\citep{cmc}. Regarding the mechanisms to maintain negative features, some methods~\citep{instdisc,pirl} keep tracking the features of all images, some directly utilize the samples within the  minibatch~\citep{simclr,cmc,clusterspread}, and \citet{moco} proposes to use a momentum encoder. \citet{byol} recently proposes to only use positive examples without negatives. Recently, \citet{pcl} argues that the lack of semantic structure is one fundamental weakness of instance discrimination, and proposes to generate prototypes by k-means clustering. However, the computational overhead and the degeneration introduced by clustering are difficult to address. \citet{debiased} also points out the possible sampling bias of instance discrimination, and proposes a debiased objective.
\vspace{-0.3cm}
\paragraph{Consistency Regularization} Consistency regularization is an important component of many successful semi-supervised learning methods. It is firstly proposed in \citet{cr}, encouraging similar predictions on perturbed versions of the same image. Several works made improvements on the way of perturbation, including using an adversarial transformation~\citep{vat}, using the prediction of a moving average or previous model~\citep{meanteacher,pi}, and using strong data augmentation~\citep{uda}. Recently, several larger pipelines are proposed~\citep{mixmatch,remixmatch,fixmatch}, in which consistency regularization still serves as a core component.
\section{Discussion}

%The topic of unsupervised visual representation learning has recently shown encouraging progress, thanks to the introduction of the contrastive learning framework. However, we argue that the instance discrimination task used by mainstream unsupervised contrastive learning methods causes ignorance of the heterogeneous similarities between image crops. In this paper, we address this issue with a consistency regularization term on the similarities between crops, inspired by some semi-supervised learning which use consistency regularization on unlabeled data. Though simple, the proposed CO2 consistently provides gains on various benchmarks. 

Unsupervised visual representation learning has shown encouraging progress recently, thanks to the introduction of instance discrimination and the contrastive learning framework. However, in this paper, we point out that instance discrimination is ignorant of the heterogeneous similarities between image crops. We address this issue with a consistency regularization term on the similarities between crops, inspired by semi-supervised learning methods which impose consistency regularization on unlabeled data. In such a simple way, the proposed CO2 consistently improves on supervised and semi-supervised image classification. It also transfers to other datasets and downstream tasks.

More broadly, we encourage researchers to rethink label correctness in existing pretext tasks. Taking instance discrimination as an example, we show that a pretext task itself could be, in fact, a semi-supervised learning task. It might be harmful to think of the pretext task as a simple pure supervised task by assuming the unknown labels are negatives. In addition, our work relaxes the stereotype restriction that pretext task labels should always be known and clean. We hope this relaxation can give rise to novel pretext tasks which exploit noisy labels or partially-available labels, making a better usage of the data without human annotation.

\bibliography{main}

\begin{thebibliography}{57}
\providecommand{\natexlab}[1]{#1}
\providecommand{\url}[1]{\texttt{#1}}
\expandafter\ifx\csname urlstyle\endcsname\relax
  \providecommand{\doi}[1]{doi: #1}\else
  \providecommand{\doi}{doi: \begingroup \urlstyle{rm}\Url}\fi

\bibitem[Asano et~al.(2019)Asano, Rupprecht, and Vedaldi]{selflabel}
YM~Asano, C~Rupprecht, and A~Vedaldi.
\newblock Self-labelling via simultaneous clustering and representation
  learning.
\newblock In \emph{ICLR}, 2019.

\bibitem[Bachman et~al.(2019)Bachman, Hjelm, and Buchwalter]{amdim}
Philip Bachman, R~Devon Hjelm, and William Buchwalter.
\newblock Learning representations by maximizing mutual information across
  views.
\newblock In \emph{NeurIPS}, 2019.

\bibitem[Berthelot et~al.(2019{\natexlab{a}})Berthelot, Carlini, Cubuk,
  Kurakin, Sohn, Zhang, and Raffel]{remixmatch}
David Berthelot, Nicholas Carlini, Ekin~D Cubuk, Alex Kurakin, Kihyuk Sohn, Han
  Zhang, and Colin Raffel.
\newblock Remixmatch: Semi-supervised learning with distribution matching and
  augmentation anchoring.
\newblock In \emph{ICLR}, 2019{\natexlab{a}}.

\bibitem[Berthelot et~al.(2019{\natexlab{b}})Berthelot, Carlini, Goodfellow,
  Papernot, Oliver, and Raffel]{mixmatch}
David Berthelot, Nicholas Carlini, Ian Goodfellow, Nicolas Papernot, Avital
  Oliver, and Colin~A Raffel.
\newblock Mixmatch: A holistic approach to semi-supervised learning.
\newblock In \emph{NeurIPS}, 2019{\natexlab{b}}.

\bibitem[Boser et~al.(1992)Boser, Guyon, and Vapnik]{svm}
Bernhard~E Boser, Isabelle~M Guyon, and Vladimir~N Vapnik.
\newblock A training algorithm for optimal margin classifiers.
\newblock 1992.

\bibitem[Caron et~al.(2018)Caron, Bojanowski, Joulin, and
  Douze]{deepclustering}
Mathilde Caron, Piotr Bojanowski, Armand Joulin, and Matthijs Douze.
\newblock Deep clustering for unsupervised learning of visual features.
\newblock In \emph{ECCV}, 2018.

\bibitem[Chen et~al.(2017)Chen, Papandreou, Kokkinos, Murphy, and
  Yuille]{deeplab}
Liang-Chieh Chen, George Papandreou, Iasonas Kokkinos, Kevin Murphy, and Alan~L
  Yuille.
\newblock Deeplab: Semantic image segmentation with deep convolutional nets,
  atrous convolution, and fully connected crfs.
\newblock \emph{TPAMI}, 2017.

\bibitem[Chen et~al.(2020{\natexlab{a}})Chen, Kornblith, Norouzi, and
  Hinton]{simclr}
Ting Chen, Simon Kornblith, Mohammad Norouzi, and Geoffrey Hinton.
\newblock A simple framework for contrastive learning of visual
  representations.
\newblock \emph{arXiv preprint arXiv:2002.05709}, 2020{\natexlab{a}}.

\bibitem[Chen et~al.(2020{\natexlab{b}})Chen, Fan, Girshick, and He]{mocov2}
Xinlei Chen, Haoqi Fan, Ross Girshick, and Kaiming He.
\newblock Improved baselines with momentum contrastive learning.
\newblock \emph{arXiv preprint arXiv:2003.04297}, 2020{\natexlab{b}}.

\bibitem[Chuang et~al.(2020)Chuang, Robinson, Yen-Chen, Torralba, and
  Jegelka]{debiased}
Ching-Yao Chuang, Joshua Robinson, Lin Yen-Chen, Antonio Torralba, and Stefanie
  Jegelka.
\newblock Debiased contrastive learning.
\newblock \emph{arXiv preprint arXiv:2007.00224}, 2020.

\bibitem[Deng et~al.(2009)Deng, Dong, Socher, Li, Li, and Fei-Fei]{imagenet}
Jia Deng, Wei Dong, Richard Socher, Li-Jia Li, Kai Li, and Li~Fei-Fei.
\newblock Imagenet: A large-scale hierarchical image database.
\newblock In \emph{CVPR}, 2009.

\bibitem[Doersch \& Zisserman(2017)Doersch and Zisserman]{multi}
Carl Doersch and Andrew Zisserman.
\newblock Multi-task self-supervised visual learning.
\newblock In \emph{ICCV}, 2017.

\bibitem[Doersch et~al.(2015)Doersch, Gupta, and Efros]{contextprediction}
Carl Doersch, Abhinav Gupta, and Alexei~A Efros.
\newblock Unsupervised visual representation learning by context prediction.
\newblock In \emph{ECCV}, 2015.

\bibitem[Donahue \& Simonyan(2019)Donahue and Simonyan]{bigbigan}
Jeff Donahue and Karen Simonyan.
\newblock Large scale adversarial representation learning.
\newblock In \emph{NeurIPS}, 2019.

\bibitem[Donahue et~al.(2016)Donahue, Kr{\"a}henb{\"u}hl, and Darrell]{bigan}
Jeff Donahue, Philipp Kr{\"a}henb{\"u}hl, and Trevor Darrell.
\newblock Adversarial feature learning.
\newblock In \emph{ICLR}, 2016.

\bibitem[Dosovitskiy et~al.(2014)Dosovitskiy, Springenberg, Riedmiller, and
  Brox]{exemplar}
Alexey Dosovitskiy, Jost~Tobias Springenberg, Martin Riedmiller, and Thomas
  Brox.
\newblock Discriminative unsupervised feature learning with convolutional
  neural networks.
\newblock In \emph{NeurIPS}, 2014.

\bibitem[Everingham et~al.(2015)Everingham, Eslami, Van~Gool, Williams, Winn,
  and Zisserman]{pascal}
Mark Everingham, SM~Ali Eslami, Luc Van~Gool, Christopher~KI Williams, John
  Winn, and Andrew Zisserman.
\newblock The pascal visual object classes challenge: A retrospective.
\newblock \emph{IJCV}, 2015.

\bibitem[Gidaris et~al.(2018)Gidaris, Singh, and Komodakis]{rotation}
Spyros Gidaris, Praveer Singh, and Nikos Komodakis.
\newblock Unsupervised representation learning by predicting image rotations.
\newblock In \emph{ICLR}, 2018.

\bibitem[Goyal et~al.(2019)Goyal, Mahajan, Gupta, and Misra]{fairssl}
Priya Goyal, Dhruv Mahajan, Abhinav Gupta, and Ishan Misra.
\newblock Scaling and benchmarking self-supervised visual representation
  learning.
\newblock In \emph{ICCV}, 2019.

\bibitem[Grill et~al.(2020)Grill, Strub, Altch{\'e}, Tallec, Richemond,
  Buchatskaya, Doersch, Pires, Guo, Azar, et~al.]{byol}
Jean-Bastien Grill, Florian Strub, Florent Altch{\'e}, Corentin Tallec,
  Pierre~H Richemond, Elena Buchatskaya, Carl Doersch, Bernardo~Avila Pires,
  Zhaohan~Daniel Guo, Mohammad~Gheshlaghi Azar, et~al.
\newblock Bootstrap your own latent: A new approach to self-supervised
  learning.
\newblock \emph{arXiv preprint arXiv:2006.07733}, 2020.

\bibitem[Hadsell et~al.(2006)Hadsell, Chopra, and LeCun]{contrasiveloss}
Raia Hadsell, Sumit Chopra, and Yann LeCun.
\newblock Dimensionality reduction by learning an invariant mapping.
\newblock In \emph{CVPR}, 2006.

\bibitem[Hariharan et~al.(2011)Hariharan, Arbel{\'a}ez, Bourdev, Maji, and
  Malik]{vocaug}
Bharath Hariharan, Pablo Arbel{\'a}ez, Lubomir Bourdev, Subhransu Maji, and
  Jitendra Malik.
\newblock Semantic contours from inverse detectors.
\newblock In \emph{ICCV}, 2011.

\bibitem[He et~al.(2016)He, Zhang, Ren, and Sun]{resnet}
Kaiming He, Xiangyu Zhang, Shaoqing Ren, and Jian Sun.
\newblock Deep residual learning for image recognition.
\newblock In \emph{CVPR}, 2016.

\bibitem[He et~al.(2017)He, Gkioxari, Doll{\'a}r, and Girshick]{maskrcnn}
Kaiming He, Georgia Gkioxari, Piotr Doll{\'a}r, and Ross Girshick.
\newblock Mask r-cnn.
\newblock In \emph{ICCV}, 2017.

\bibitem[He et~al.(2020)He, Fan, Wu, Xie, and Girshick]{moco}
Kaiming He, Haoqi Fan, Yuxin Wu, Saining Xie, and Ross Girshick.
\newblock Momentum contrast for unsupervised visual representation learning.
\newblock In \emph{CVPR}, 2020.

\bibitem[H{\'e}naff et~al.(2019)H{\'e}naff, Razavi, Doersch, Eslami, and
  Oord]{cpcv2}
Olivier~J H{\'e}naff, Ali Razavi, Carl Doersch, SM~Eslami, and Aaron van~den
  Oord.
\newblock Data-efficient image recognition with contrastive predictive coding.
\newblock \emph{arXiv preprint arXiv:1905.09272}, 2019.

\bibitem[Hinton et~al.(2015)Hinton, Vinyals, and Dean]{knowledgedistillation}
Geoffrey Hinton, Oriol Vinyals, and Jeff Dean.
\newblock Distilling the knowledge in a neural network.
\newblock In \emph{NeurIPS Deep Learning and Representation Learning Workshop},
  2015.

\bibitem[Hjelm et~al.(2018)Hjelm, Fedorov, Lavoie-Marchildon, Grewal, Bachman,
  Trischler, and Bengio]{cpc}
R~Devon Hjelm, Alex Fedorov, Samuel Lavoie-Marchildon, Karan Grewal, Phil
  Bachman, Adam Trischler, and Yoshua Bengio.
\newblock Learning deep representations by mutual information estimation and
  maximization.
\newblock \emph{arXiv preprint arXiv:1808.06670}, 2018.

\bibitem[Hjelm et~al.(2019)Hjelm, Fedorov, Lavoie-Marchildon, Grewal, Bachman,
  Trischler, and Bengio]{dim}
R~Devon Hjelm, Alefx Fedorov, Samuel Lavoie-Marchildon, Karan Grewal, Phil
  Bachman, Adam Trischler, and Yoshua Bengio.
\newblock Learning deep representations by mutual information estimation and
  maximization.
\newblock In \emph{ICLR}, 2019.

\bibitem[Ioffe \& Szegedy(2015)Ioffe and Szegedy]{bn}
Sergey Ioffe and Christian Szegedy.
\newblock Batch normalization: Accelerating deep network training by reducing
  internal covariate shift.
\newblock In \emph{ICML}, 2015.

\bibitem[Khosla et~al.(2020)Khosla, Teterwak, Wang, Sarna, Tian, Isola,
  Maschinot, Liu, and Krishnan]{supervisedcl}
Prannay Khosla, Piotr Teterwak, Chen Wang, Aaron Sarna, Yonglong Tian, Phillip
  Isola, Aaron Maschinot, Ce~Liu, and Dilip Krishnan.
\newblock Supervised contrastive learning.
\newblock \emph{arXiv preprint arXiv:2004.11362}, 2020.

\bibitem[Kolesnikov et~al.(2019)Kolesnikov, Zhai, and Beyer]{revisting}
Alexander Kolesnikov, Xiaohua Zhai, and Lucas Beyer.
\newblock Revisiting self-supervised visual representation learning.
\newblock In \emph{CVPR}, 2019.

\bibitem[Laine \& Aila(2017)Laine and Aila]{pi}
Samuli Laine and Timo Aila.
\newblock Temporal ensembling for semi-supervised learning.
\newblock In \emph{ICLR}, 2017.

\bibitem[Larsson et~al.(2017)Larsson, Maire, and
  Shakhnarovich]{colorizationproxy}
G.~Larsson, M.~Maire, and G.~Shakhnarovich.
\newblock Colorization as a proxy task for visual understanding.
\newblock In \emph{CVPR}, 2017.

\bibitem[Li et~al.(2020)Li, Zhou, Xiong, Socher, and Hoi]{pcl}
Junnan Li, Pan Zhou, Caiming Xiong, Richard Socher, and Steven~CH Hoi.
\newblock Prototypical contrastive learning of unsupervised representations.
\newblock \emph{arXiv preprint arXiv:2005.04966}, 2020.

\bibitem[Long et~al.(2015)Long, Shelhamer, and Darrell]{fcn}
Jonathan Long, Evan Shelhamer, and Trevor Darrell.
\newblock Fully convolutional networks for semantic segmentation.
\newblock In \emph{CVPR}, 2015.

\bibitem[Loshchilov \& Hutter(2016)Loshchilov and Hutter]{cosinedecay}
Ilya Loshchilov and Frank Hutter.
\newblock Sgdr: Stochastic gradient descent with warm restarts.
\newblock In \emph{ICLR}, 2016.

\bibitem[Misra \& van~der Maaten(2020)Misra and van~der Maaten]{pirl}
Ishan Misra and Laurens van~der Maaten.
\newblock Self-supervised learning of pretext-invariant representations.
\newblock In \emph{CVPR}, 2020.

\bibitem[Miyato et~al.(2018)Miyato, Maeda, Koyama, and Ishii]{vat}
Takeru Miyato, Shin-ichi Maeda, Masanori Koyama, and Shin Ishii.
\newblock Virtual adversarial training: a regularization method for supervised
  and semi-supervised learning.
\newblock \emph{TPAMI}, 2018.

\bibitem[Noroozi \& Favaro(2016)Noroozi and Favaro]{jigsaw}
Mehdi Noroozi and Paolo Favaro.
\newblock Unsupervised learning of visual representations by solving jigsaw
  puzzles.
\newblock In \emph{ECCV}, 2016.

\bibitem[Noroozi et~al.(2017)Noroozi, Pirsiavash, and Favaro]{counting}
Mehdi Noroozi, Hamed Pirsiavash, and Paolo Favaro.
\newblock Representation learning by learning to count.
\newblock In \emph{ICCV}, 2017.

\bibitem[Pathak et~al.(2016)Pathak, Krahenbuhl, Donahue, Darrell, and
  Efros]{inpainting}
Deepak Pathak, Philipp Krahenbuhl, Jeff Donahue, Trevor Darrell, and Alexei~A
  Efros.
\newblock Context encoders: Feature learning by inpainting.
\newblock In \emph{CVPR}, 2016.

\bibitem[Ren et~al.(2015)Ren, He, Girshick, and Sun]{fasterrcnn}
Shaoqing Ren, Kaiming He, Ross Girshick, and Jian Sun.
\newblock Faster r-cnn: Towards real-time object detection with region proposal
  networks.
\newblock In \emph{NeurIPS}, 2015.

\bibitem[Sajjadi et~al.(2016)Sajjadi, Javanmardi, and Tasdizen]{cr}
Mehdi Sajjadi, Mehran Javanmardi, and Tolga Tasdizen.
\newblock Regularization with stochastic transformations and perturbations for
  deep semi-supervised learning.
\newblock In \emph{NeurIPS}, 2016.

\bibitem[Saunshi et~al.(2019)Saunshi, Plevrakis, Arora, Khodak, and
  Khandeparkar]{theoretical}
Nikunj Saunshi, Orestis Plevrakis, Sanjeev Arora, Mikhail Khodak, and
  Hrishikesh Khandeparkar.
\newblock A theoretical analysis of contrastive unsupervised representation
  learning.
\newblock In \emph{ICML}, 2019.

\bibitem[Sohn et~al.(2020)Sohn, Berthelot, Li, Zhang, Carlini, Cubuk, Kurakin,
  Zhang, and Raffel]{fixmatch}
Kihyuk Sohn, David Berthelot, Chun-Liang Li, Zizhao Zhang, Nicholas Carlini,
  Ekin~D Cubuk, Alex Kurakin, Han Zhang, and Colin Raffel.
\newblock Fixmatch: Simplifying semi-supervised learning with consistency and
  confidence.
\newblock In \emph{NeurIPS}, 2020.

\bibitem[Szegedy et~al.(2016)Szegedy, Vanhoucke, Ioffe, Shlens, and Wojna]{ls}
Christian Szegedy, Vincent Vanhoucke, Sergey Ioffe, Jon Shlens, and Zbigniew
  Wojna.
\newblock Rethinking the inception architecture for computer vision.
\newblock In \emph{CVPR}, 2016.

\bibitem[Tarvainen \& Valpola(2017)Tarvainen and Valpola]{meanteacher}
Antti Tarvainen and Harri Valpola.
\newblock Mean teachers are better role models: Weight-averaged consistency
  targets improve semi-supervised deep learning results.
\newblock In \emph{NeurIPS}, 2017.

\bibitem[Tian et~al.(2019)Tian, Krishnan, and Isola]{cmc}
Yonglong Tian, Dilip Krishnan, and Phillip Isola.
\newblock Contrastive multiview coding.
\newblock \emph{arXiv preprint arXiv:1906.05849}, 2019.

\bibitem[Wei et~al.(2019)Wei, Xie, Ren, Xia, Su, Liu, Tian, and
  Yuille]{arbitrarypuzzle}
Chen Wei, Lingxi Xie, Xutong Ren, Yingda Xia, Chi Su, Jiaying Liu, Qi~Tian, and
  Alan~L Yuille.
\newblock Iterative reorganization with weak spatial constraints: Solving
  arbitrary jigsaw puzzles for unsupervised representation learning.
\newblock In \emph{CVPR}, 2019.

\bibitem[Wu et~al.(2019)Wu, Kirillov, Massa, Lo, and Girshick]{detectron2}
Yuxin Wu, Alexander Kirillov, Francisco Massa, Wan-Yen Lo, and Ross Girshick.
\newblock Detectron2.
\newblock \url{https://github.com/facebookresearch/detectron2}, 2019.

\bibitem[Wu et~al.(2018)Wu, Xiong, Yu, and Lin]{instdisc}
Zhirong Wu, Yuanjun Xiong, Stella~X Yu, and Dahua Lin.
\newblock Unsupervised feature learning via non-parametric instance
  discrimination.
\newblock In \emph{CVPR}, 2018.

\bibitem[Xie et~al.(2019)Xie, Dai, Hovy, Luong, and Le]{uda}
Qizhe Xie, Zihang Dai, Eduard Hovy, Minh-Thang Luong, and Quoc~V Le.
\newblock Unsupervised data augmentation for consistency training.
\newblock \emph{arXiv preprint arXiv:1904.12848}, 2019.

\bibitem[Ye et~al.(2019)Ye, Zhang, Yuen, and Chang]{clusterspread}
Mang Ye, Xu~Zhang, Pong~C Yuen, and Shih-Fu Chang.
\newblock Unsupervised embedding learning via invariant and spreading instance
  feature.
\newblock In \emph{CVPR}, 2019.

\bibitem[Zhang et~al.(2016)Zhang, Isola, and Efros]{colorization}
Richard Zhang, Phillip Isola, and Alexei~A Efros.
\newblock Colorful image colorization.
\newblock In \emph{ECCV}, 2016.

\bibitem[Zhong et~al.(2020)Zhong, Chen, Jin, and Hua]{deeprobustclustering}
Huasong Zhong, Chong Chen, Zhongming Jin, and Xian-Sheng Hua.
\newblock Deep robust clustering by contrastive learning.
\newblock \emph{arXiv preprint arXiv:2008.03030}, 2020.

\bibitem[Zhuang et~al.(2019)Zhuang, Zhai, and Yamins]{la}
Chengxu Zhuang, Alex~Lin Zhai, and Daniel Yamins.
\newblock Local aggregation for unsupervised learning of visual embeddings.
\newblock In \emph{CVPR}, 2019.

\end{thebibliography}
\bibliographystyle{iclr2021_conference}

\newpage
\appendix
\section{Appendix}
\subsection{Implementation Details of Contrastive Pre-Training}
We evaluate our approach based on MoCo~\citep{moco}. MoCo has two different encoders to encode queries and keys respectively. The query encoder is updated with respect to the loss function, while the key encoder is an exponential moving average of the query encoder. The keys are stored in a dynamic memory bank, whose entries are updated at every training step with the current mini-batch enqueued and the oldest mini-batch dequeued. The backbone is a standard ResNet-50~\citep{resnet}, and features after the global average pooling layer are projected to 128-D vectors~\citep{instdisc}, normalized by $\ell_2$ norm. The size of the memory bank (i.e., the number of negative samples) is 65,536 and the momentum to update the key encoder is 0.999. $\tau_{ins}$ is 0.07 for MoCo variants and 0.2 for MoCo v2 variants, which are the default settings of these two methods.

We use momentum SGD with momentum 0.9 and weight decay 1e-4. The batch size is 256 on 4 GPUs. To prevent potential information leak with Batch Normalization (BN)~\citep{bn}, shuffling BN~\citep{moco} is performed. The model is trained for 200 epochs with the initial learning rate of 0.03. The learning rate is multiplied by 0.1 after 120 and 160 epochs for MoCo v1, while cosine decayed~\citep{cosinedecay} for MoCo v2. We keep aligned all training details with MoCo except the number of GPUs. This could be problematic since it changes the per-worker minibatch size, which is related to potential information leaks pointed by ~\citet{moco}. However, we do not notice much difference when reproducing MoCo with 4 GPUs. Our reproduced MoCo v2 with 4 GPUs reaches the  accuracy of 67.6\% on the linear classification protocol, 0.1\% higher than 67.5\% reported in its paper. For the hyper-parameters of the proposed consistency term, we set $\tau_{con}s$ as 0.04 and $\alpha$ as 10 for the MoCo v1-based CO2, and $\tau_{con}$ as 0.05, $\alpha$ as 0.3 for the MoCo v2-based variant.

\subsection{Implemetation Details of Downstream Tasks}
\paragraph{Linear Classification} We freeze the backbone network including the batch normalization parameters, and train a linear classifier consisting of a fully-connected layer followed by softmax on the 2048-D features following the global average pooling layer. We train for 100 epochs. The learning rate is initialized as 15 and decayed by 0.1 every 20 epoch after the first 60 epochs. We set weight decay as 0 and momentum as 0.9. Only random cropping with random horizontal flipping is used as data augmentation. 
\vspace{-0.3cm}
\paragraph{Semi-Supervised Learning} We finetune the pre-trained model for 20 epochs with learning rate starting from 0.01 for the base model and 1.0 for the randomly initialized classification head, decayed by 0.2 after 12 and 16 epochs. Momentum is set to 0.9. Weight decay is 5e-4 for MoCo v1 and 1e-4 for MoCo v2. Only random cropping with random horizontal flipping is used as data augmentation.
\vspace{-0.3cm}
\paragraph{Classification on PASCAL VOC} Following  the  evaluation  setup  in  \citet{fairssl},  we  train a  linearSVM~\citep{svm} on the frozen 2048-D features extracted after the global average pooling layer. The models are trained on \texttt{trainval2007} split and tested on \texttt{test2007}. The hyper-parameters are selected based on a held-out subset of the training set. 
\vspace{-0.3cm}
\paragraph{Detection on PASCAL VOC} Following the detection benchmark set up in \citet{moco}, we use FasterR-CNN~\citep{fasterrcnn} object detector and ResNet-50 C4~\citep{maskrcnn} backbone, implemented in Detectron2~\citep{detectron2}. We finetune all the layers including the batch normalization parameters for 24k iterations on the \texttt{trainval07+12} split and test on \texttt{test2007} set. The hyper-parameters are the same as the counterpart with supervised ImageNet initialization and MoCo. To calibrate the small feature magnitude due to the output normalization in the unsupervised pre-training stage, two extra batch normalization layers are introduced, one is followed by the regional proposal head whose gradients are divided by 10 and the other is followed by the box prediction head.
\vspace{-0.3cm}
\paragraph{Segmentation on PASCAL VOC} Following the setup in \citet{moco}, an FCN-based~\citep{fcn} architecture with atrous convolutions~\citep{deeplab} is used and ResNet-50 is the backbone. The training set is \texttt{train\_aug2012}~\citep{vocaug} and the testing set is \texttt{val2012}. Initialized with CO2 models, we finetune all layers for 50 epochs (~33k iterations) with batch size 16, initial learning rate 0.003, weight decay 1e-4 and momentum 0.9.

\end{document}